\pdfoutput=1

\documentclass[11pt]{article}

\usepackage{acl}

\usepackage{times}
\usepackage{latexsym}
\usepackage{graphicx}
\usepackage{amsmath}
\usepackage{amssymb}
\usepackage{cleveref}
\newcommand{\Scref}[1]{\S\ref{#1}}
\usepackage{booktabs} 
\usepackage{multicol}
\usepackage{natbib}
\usepackage{enumitem}

\usepackage[T1]{fontenc}

\usepackage[utf8]{inputenc}

\usepackage{microtype}

\usepackage{inconsolata}

\title{How does Multi-Task Training Affect Transformer In-Context Capabilities? Investigations with Function Classes}

\author{Harmon Bhasin$^1$,
  Timothy Ossowski$^1$,
  Yiqiao Zhong$^1$,
  Junjie Hu$^1$ \\
  $^1$University of Wisconsin, Madison, WI, USA \\
  \texttt{hsbhasin@wisc.edu}, \texttt{ossowski@wisc.edu}, \texttt{yiqiao.zhong@wisc.edu}, \texttt{junjie.hu@wisc.edu}}

\begin{document}
\maketitle

\begin{abstract}
Large language models (LLM) have recently shown the extraordinary ability to perform unseen tasks based on few-shot examples provided as text, also known as in-context learning (ICL). While recent works have attempted to understand the mechanisms driving ICL, few have explored training strategies that incentivize these models to generalize to multiple tasks. Multi-task learning (MTL) for generalist models is a promising direction that offers transfer learning potential, enabling large parameterized models to be trained from simpler, related tasks. In this work, we investigate the combination of MTL with ICL to build models that efficiently learn tasks while being robust to out-of-distribution examples. We propose several effective curriculum learning strategies that allow ICL models to achieve higher data efficiency and more stable convergence. Our experiments\footnote{Our code and models are available at \url{https://github.com/harmonbhasin/curriculum_learning_icl}} reveal that ICL models can effectively learn difficult tasks by training on progressively harder tasks while mixing in prior tasks, denoted as mixed curriculum in this work.

\end{abstract}

\section{Introduction} \label{introduction}

Recently, the emergence of in-context-learning capabilities in LLMs has revolutionized the field of NLP \cite{wei2022emergent}. By pre-training with next-word predictions, these models can be prompted with few-shot examples and make accurate in-context predictions during inference \cite{brown2020language}. The ICL capability demonstrated even by smaller Transformer models presents an alternative way to understand LLMs \cite{dong2023survey,pmlr-v202-li23l,lu2023emergent}. To empirically understand this phenomenon, \citet{garg2022icl} focus on learning a single function class in-context by a Transformer model. Their model achieves competitive normalized mean-squared error (MSE) compared to the optimal ordinary least squares estimator when performing in-context linear regression. Nevertheless, these models sometimes fail to converge and often struggle to generalize to more challenging function classes. 
While the follow-up studies~\cite{akyurek2023what,pmlr-v202-von-oswald23a,yang2023iterative} have extensively analyzed how these models conduct ICL, little work exists exploring 
how training on \textit{multiple function classes} can enable Transformer models to generalize and perform ICL more efficiently. As we believe that these generalist models are designed to perform multiple tasks, there is a need to study the multi-task ICL capability of these models, which is missing in the literature.


From prior multi-task learning studies~\cite{zhang2023survey,ruder2017overview,weiss.etal_2016}, models can be trained on multiple related tasks to improve their performance on individual tasks. Despite its popularity, MTL has been difficult to understand in Transformer models when trained on natural language, most likely due to the difficulty of ranking and scheduling language tasks \cite{crawshaw2020multitask}. 
However, the newly introduced framework of learning function classes in context~\cite{garg2022icl} provides an easier way to study this MTL paradigm. For example, the difficulty of a polynomial function class can be scaled by changing its degree (e.g., linear to quadratic), or changing the input distribution (e.g., Gaussian to Gaussian with decaying eigenvalues). Motivated by this new paradigm, we conduct a systematic analysis by training a Transformer on varying function class families and input distributions in a multi-task manner to examine if the same principles from MTL carry over into ICL.

During training, we explore different curriculum learning strategies to schedule the ICL tasks of multiple function classes: \textit{sequential}, \textit{mixed}, and \textit{random} (\Scref{curriculum_learning}). For benchmarking, we train another set of models only on a single function class family following~\citet{garg2022icl}. We quantitatively and qualitatively compare our models trained with and without curriculum across all tasks and analyze the normalized MSE and attention matrices (\Scref{results}). 
Our experiments show that curriculum learning is more data-efficient, achieving comparable performance to single-task models using only 1/9 of the training data. These curriculum models can also obtain an optimal MSE in function classes where none of the single-task models converge. 

\section{Related Work}

\paragraph{In-context Learning}

In-context learning has garnered increasing attention in the past few years \cite{dong2023survey}, and many papers have analyzed ICL with natural language \cite{min-etal-2022-rethinking,xie2022an,min-etal-2022-metaicl}. It was not until \citet{garg2022icl} that the analysis of ICL through the paradigm of function class learning emerged. \citet{garg2022icl} showed that Transformers can learn linear regression close to the optimal ordinary least squares estimator, and other more complex function classes with respectable accuracy. However, they found that some function classes (e.g. Gaussian with decaying eigenvalues) were hard to learn by Transformers, as the training loss failed to converge. \citet{yadlowsky2023pretraining} investigated a framework similar to \citet{garg2022icl}, where they explored training models on a mixture of function classes; however, they did not delve into curriculum learning strategies. Many papers also explored how ICL works, with current literature pointing to it being a fuzzy gradient descent~\cite{akyurek2023what,pmlr-v202-von-oswald23a,yang2023iterative}. Additional theoretical work has examined how transformers can implement near-optimal regression algorithms and has analyzed stability conditions for ICL \cite{li2023transformers}.

\paragraph{Curriculum Learning} 
\citet{bengio.etal_2009} first introduced curriculum learning as a way to train models similar to the way that humans learned, by learning tasks in order from easy to hard. This work inspired a new area of research focused on utilizing curriculum learning in different contexts \cite{xu.etal_2020,wang.etal_2021,soviany.etal_2022}. Among this exploration has been more complex curriculum learning strategies in well studied contexts \cite{graves.etal_2017,varshney.etal_2022}. The novel function learning problem formulation in \citet{garg2022icl} has encouraged us to focus on simple curriculum learning strategies that have been well-studied. Our sequential curriculum aligns with the definition provided by \citet{bengio.etal_2009}. We have adapted our mixed curriculum from the standard curriculum, which is referred to as a ``balanced curriculum'' in \citet{soviany.etal_2022}. Finally, our random curriculum serves as a baseline approach, as described in \citet{soviany.etal_2022}. By conducting the first exploratory study on these simple, widely-used curriculum learning strategies, we pave the way for more sophisticated strategies. 

\paragraph{Attention Analysis}

Transformers~\cite{vaswani2017attention} have revolutionized our capabilities of performing tasks in a variety of fields. Recognizing the significance of attention behind Transformers, we aimed to analyze it in the context of ICL, akin to previous work~\cite{clark-etal-2019-bert}. \citet{olsson2022context} and \citet{elhage2021mathematical} found that specific attention heads, specified as ``induction heads'', were responsible for the ICL ability of Transformers, both in large and small Transformers. To measure this, they created their own metric. Intrigued by the possibility that certain heads might attend to specific tasks within a multi-task framework, we decided to visualize the attention matrix. Inspired by \citet{vig-belinkov-2019-analyzing} that showed a simple and interpretable way to visualize attention, we used this approach as a proxy to develop our own analyses of the attention matrices in this study. Furthermore, other recent studies have focused on summarizing attention flow through Transformer models from input embeddings to later layers with attention rollout \cite{abnar2020quantifying}.

\paragraph{Instruction Prompting}

Instruction prompting has been widely used in natural language tasks to improve accuracy and tends to be robust to variations during test time \cite{liu2023prompt}. \citet{wei2023larger} showed that models of different architectures responded differently to instruction tokens, with the format of the instruction affecting multi-task settings. \citet{yin-etal-2023-read} showed that providing key information in tasks in a common format improved the ability of the model to learn the task. Recently frameworks have emerged that prompt LLMs with intermediate reasoning steps to elicit better reasoning capabilities \cite{wei2022chain}, known as Chain of Thought (CoT) prompting. \cite{besta2023graph} and \cite{yao2023tree} extend CoT prompting to consider multiple reasoning paths to improve performance. Future work may consider using these methods to improve ICL in the multi-task setting.

\section{Methods} \label{methods}


\subsection{Problem Definition} \label{overview}
Following \citet{garg2022icl}, we define the problem of ICL as passing in an $i$-shot sequence $S^i=(x_1,f(x_1), x_2, f(x_2), \dots, x_n, f(x_i), x_{i+1})$ to the Transformer (denoted as $M_\theta$) and generating an output $M_\theta(S^i)$ to predict the ground-truth $f(x_{i+1})$, where the examples have not been seen during training. We refer to this $i$-shot prediction problem, where input is given in pairs, as ICL. 

We consider a data-generating process where $d$-dimensional inputs are drawn from any arbitrary distribution (i.e., $x_i \sim \mathcal{D}_x$) and a function $f$ is sampled from the class of functions related to single-index probabilist's normalized Hermite polynomials (i.e, $f\sim \mathcal{F}$). 

Similar to~\citet{garg2022icl}, the training objective is to minimize the squared error $l(\cdot,\cdot)$ between the prediction $M_\theta (S^i)$ and ground-truth $f(x_{i+1})$: 
\vspace{-0.2cm}
\begin{align} \nonumber
    \mathop{min}_{\theta} \mathbb{ E}_{S^i} \left[\frac{1}{k+1} \sum^{k}_{i=0} l (M_\theta (S^i),f(x_{i+1})) \right].
\end{align}
Appendix~\ref{experimental_settings} shows more training details.

\subsection{Tasks} \label{tasks}

We explore two types of tasks: learning a function class and learning a data distribution (see Appendix \ref{distribution_learning}). We consider a single-index function: 
\vspace{-0.2cm}
\begin{align} \nonumber
    f(x) = \varphi(\langle x, w \rangle).
\end{align}

\paragraph{Function Class Learning}

We look at the class of functions derived from normalized probabilist's Hermite polynomial with degree $n$ and constants removed, i.e., $\frac{1}{\sqrt{n!}}He_n(x)$, which satisfies orthogonality. This is useful as it guarantees that the function values of all tasks are uncorrelated. For each task, we separately sample $x$ and $w$ from an isotropic Gaussian distribution, where $w$ remains constant for an $i$-shot sequence. We define $K=3$ polynomial function classes as follows: denoting $t = \langle x, w \rangle$, we pick $\varphi \in \{\varphi_{\text{linear}}, \varphi_{\text{quadratic}}, \varphi_{\text{cubic}}\}$ for three function classes $\mathcal{F}_1$, $\mathcal{F}_2$, $\mathcal{F}_3$, respectively.

\vspace{-0.5cm}

\begin{align}\nonumber
    &\varphi_{\text{linear}}(t) = t, \\ \nonumber
    &\varphi_{\text{quadratic}}(t) = \frac{1}{\sqrt{2}}(t + \frac{1}{\sqrt{2}}(t^2 - 1)) \\ \nonumber
    &\varphi_{\text{cubic}}(t) = \frac{1}{\sqrt{3}}(t + \frac{1}{\sqrt{2}}(t^2 - 1) + \frac{1}{\sqrt{6}}(t^3 - 3t)) 
\end{align}

\subsection{Curriculum Learning} \label{curriculum_learning}
We define the total training steps to be $T$, where the $t$-th training step ranges from $t=1,2,\dots,T$. Our curriculum learning strategy (\textit{sequential}, \textit{mixed}, or \textit{random}) is used to allocate our $K$ tasks across training time. In this paper, we explore $K=3$ function classes defined earlier.

\paragraph{Sequential Curriculum}
We first separate the total training steps $T$ into $K$ partitions. Within the $k$-th partition of training steps, we train the model on learning a function from the $k$-th function class, in order of increasing difficulty:

\vspace{-0.4cm}
\begin{align}\nonumber
f \sim \begin{cases} 
      \mathcal{F}_1 & 1 \leq t < \frac{T}{3} \\
      \mathcal{F}_2 & \frac{T}{3} \leq t < \frac{2T}{3} \\
      \mathcal{F}_3 & \frac{2T}{3} \leq t < T
   \end{cases}
\end{align}

\paragraph{Mixed Curriculum}
We first separate the total training steps $T$ into $K$ partitions. Let $\xi$ be (uniformly) drawn from $\{1,2\}$ and $\zeta$ be (uniformly) drawn from $\{1, 2, 3\}$. We select tasks from the previous $k$ partitions with equal probability ($\mathbf{1}$ denotes the indicator function):

\vspace{-0.4cm}

\begin{align}\nonumber
f \sim \begin{cases} 
      \mathcal{F}_1 & 1 \leq t < \frac{T}{3} \\
      \sum_{s=1}^2 \mathbf{1}(\xi=s) \mathcal{F}_{s} & \frac{T}{3} \leq t < \frac{2T}{3} \\
      \sum_{s=1}^3 \mathbf{1}(\zeta=s) \mathcal{F}_{s}  & \frac{2T}{3} \leq t < T
   \end{cases}
\end{align}

\paragraph{Random Curriculum}

 At each training step $t$, we randomly sample from the list of $K$ tasks with equal probability:

\vspace{-0.4cm}
 
\begin{align}\nonumber
f \sim \sum_{s=1}^3 \mathbf{1}(\zeta=s) \mathcal{F}_{s}, \quad 1 \leq t < T
\end{align}

\subsection{Attention Analysis} \label{attention_analysis}

To understand how single and multi-task models learn, we analyze the Transformer's self-attention weights. Specifically, we mask out the attention matrices for each head to keep only the self-attention scores between each $f(x_i)$ token and its corresponding $x_{i}$ token. To summarize the head's inclination to attend to previous tokens, we aggregate these scores by taking the mean across all $f(x_i)$ tokens. We repeat this for all attention heads in all layers and plot the aggregated scores in a head-by-layer heatmap. We define a ``retrospective head'' as an attention head that has a lighter value in the heatmap, indicating that this specific head learns to attend to the previous input token when constructing a representation for the current token, a natural pattern that encourages understanding of the input-output pairs, i.e., $(x_i, f(x_i))$.  

\section{Results} \label{results}

\begin{figure}[ht!]
    \centering
    \includegraphics[width=.48\textwidth]{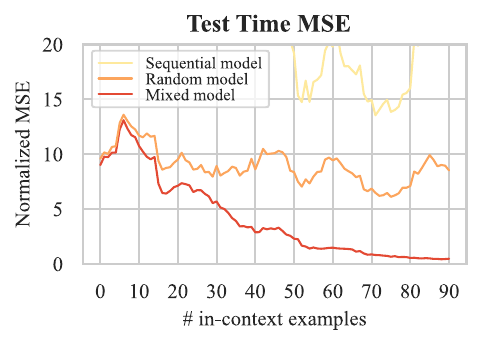}
    \vspace{-0.35in}
    \caption{Comparison of the moving average of all three curriculum learning strategies when evaluated on a quadratic function class dataset during test time. The mixed curriculum is the only model that is able to achieve an accurate normalized MSE. The random curriculum performs comparatively worse, whereas the sequential curriculum performs substantially worse (y-axis is limited in order for mixed and random curricula to be differentiated).}
    \label{fig:mixed_best_curriculum}
\end{figure}

\vspace{-5mm}
\paragraph {Curriculum Learning Comparison} \label{curriculum_comparison}

Figure~\ref{fig:mixed_best_curriculum} shows that the mixed curriculum outperforms both the random and sequential curricula when evaluating all models on a quadratic function class dataset according to mean squared errors (MSE) during test time. We find that the mixed curriculum strategy provides the most benefit towards learning multiple tasks. This is further validated in Supplementary Figure ~\ref{suppfig:model_performance_func_learning_curriculum}, which shows that the mixed curriculum is most stable over all tasks, achieving an accurate solution after sufficient few-shot examples (20/80/90-shot examples for Linear/Quadratic/Cubic respectively). We hypothesize that this is due to stable periods of training, where the model can adapt to the given function class, whereas the random curriculum does not have such a schedule. Additionally, mixed curriculum likely outperforms sequential curriculum because including tasks from previous training blocks mitigates catastrophic forgetting \cite{zhai2023investigating}. Thus, we stick with the mixed curriculum model in the following experiments.

\begin{figure*}[ht!]
    \centering
    \includegraphics[width=.99\textwidth]{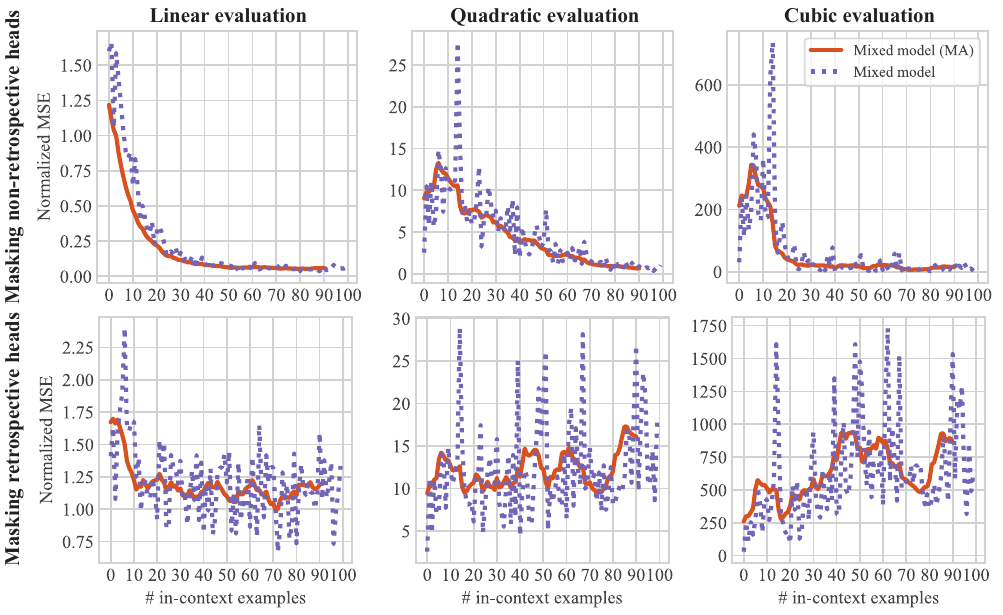}
    \vspace{-0.1in}
    \caption{Masking retrospective heads \textit{(bottom row)} causes significant increase in normalized MSE compared to non-retrospective heads \textit{(top row)} in the mixed curriculum model.}
    \label{fig:induction_heads}
\end{figure*}

\paragraph{Qualitative Attention Analysis} \label{qualitatitve_attention}

Figure~\ref{fig:induction_heads} displays how masking 7 retrospective heads (as defined in~\Scref{attention_analysis}) causes a significant increase in normalized MSE compared to 7 non-retrospective heads in the mixed curriculum model. Using our attention analysis in Supplementary Figure~\ref{suppfig:attention_analysis_func_learning_curriculum}, we identify retrospective heads as those with yellow values, whereas non-retrospective heads are highlighted with dark purple values. This supports the theory that specific heads may be reasonable for the ICL capability of these models \cite{olsson2022context}. Additionally, these retrospective heads stay the same across different task evaluations. Pairing this with the normalized MSE analysis in Supplementary Figure~\ref{suppfig:model_performance_func_learning_curriculum}, we hypothesize that these models are conducting approximations rather than learning the true tasks as the model achieves optimal, but not perfect (normalized MSE = 0) over all tasks. 



\begin{figure}[ht!]
    \includegraphics[width=.47\textwidth]{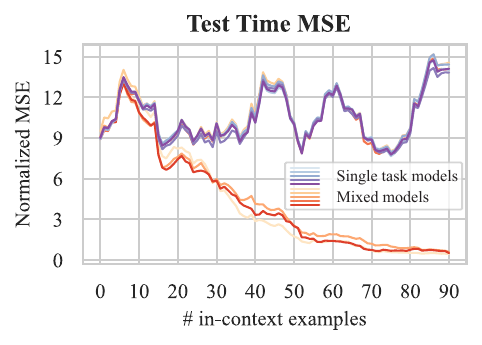}
    \vspace{-0.2in}
    \caption{Comparison of the moving average of five different seeded single-task (\textit{blue-purple}) and mixed curriculum models (\textit{orange-red}) evaluated on a quadratic function class dataset during test time. Mixed curriculum models are able to learn quadratic function classes whereas the single task models are unable to, indicated by the spikes and upward trend in normalized MSE.}
    \label{fig:curriculum_stability}
\end{figure}

\paragraph{Curriculum Learning Convergence} \label{curriculum_convergence}

Figure~\ref{fig:curriculum_stability} reveals 60\% of mixed curriculum models converge, whereas 0\% of the single-task models trained on quadratic function classes converge. Specifically, these models do not achieve optimal (below 1) normalized MSE during training time and at test time. We believe curriculum learning aids in this task, as we allow the model to warm up the training with the objective (calculate $f(x)$ from $x$) on easier tasks. In contrast, the poor performance of the single-task models may be explained by their cryptic attention patterns (Supplementary Figure~\ref{suppfig:attention_analysis_func_learning_baseline}). These findings help us understand how curriculum learning can be used to learn difficult function classes that are otherwise unlearable by single-task models.

\begin{figure}[ht!]
    \centering
    \includegraphics[width=.48\textwidth]{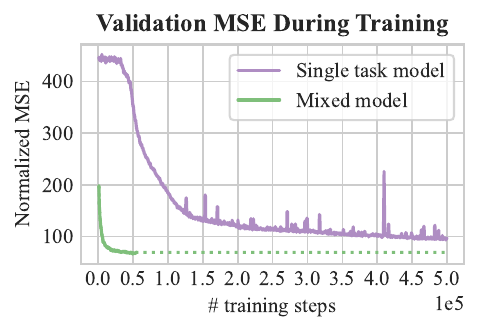}
    \vspace{-0.35in}
    \caption{Comparison of the moving average of a single-task model and a mixed curriculum model evaluated on a cubic function class dataset during training time. The mixed curriculum model is initialized with a checkpoint trained on linear and quadratic function examples, while the single-task model is initialized with random weights.}
    
    \label{fig:curriculum_efficiency}
\end{figure}

\paragraph{Curriculum Learning Data Efficiency} \label{curriculum_efficiency}

Figure~\ref{fig:curriculum_efficiency} illustrates the performance of a single-task model and a mixed curriculum model during training when evaluated on a cubic function class validation dataset. Our experiments uncover that the mixed curriculum model can improve data efficiency, learning harder tasks with fewer examples. The mixed curriculum model is pre-trained on 1/9 of the training examples seen by the single-task cubic model, yet the mixed curriculum model has better performance on the validation set. Pulling from qualitative attention analysis, we hypothesize that the mixed curriculum model is able to use its approximate understanding of the linear and quadratic function classes to improve the initial normalized MSE of a cubic function class. This explains why the cubic model starts at 450 normalized MSE, whereas the mixed model starts at 200 normalized MSE. When analyzing both models at test time (Supplementary Figure~\ref{suppfig:model_performance_func_learning_baseline} and~\ref{suppfig:model_performance_func_learning_curriculum}) the mixed model has comparable performance to the single-task cubic model. These findings suggest that curriculum learning can assist data efficiency by making use of transfer learning from easier tasks. 

\section{Discussion} \label{discussion}


In this paper, we examine how different curriculum learning strategies affect a Transformer's ICL capability. We compare these curriculum models against their respective single-task models and evaluate them across related tasks. This reveals that the mixed curriculum provides the best results, with increased data efficiency and model convergence. Our attention analysis shows that these curriculum learning models share the same retrospective heads across related tasks. Masking these retrospective heads during test time drastically drops the accuracy of these models across tasks, indicating that specific heads are responsible for ICL. This work provides the preliminary analysis necessary to explore curriculum learning in these ICL settings in natural language. We hope that these results provide an important insight into how we can better pre-train LLMs to ICL efficiently.


\section*{Limitations}
Our work investigates ICL on standard function classes which can be mathematically defined, however it may be difficult to extend our work to natural language tasks as they are hard to define. The extensibility of our work to natural language tasks therefore remains an open question. We make use of three well-known curriculum learning strategies, however, more effective strategies should be investigated. We work with a relatively small model, thus our results may not be transferable to larger models such as Llama-2 or GPT-4 and we work with noiseless data which may inflate the accuracy. Lastly, we acknowledge that ICL can be inconsistent (models only learn approximations for tasks and have varying performance across seeds) and should not be used in high-risk situations.

\bibliography{anthology,custom}

\begin{thebibliography}{40}
\expandafter\ifx\csname natexlab\endcsname\relax\def\natexlab#1{#1}\fi

\bibitem[{Abnar and Zuidema(2020)}]{abnar2020quantifying}
Samira Abnar and Willem Zuidema. 2020.
\newblock Quantifying attention flow in transformers.
\newblock \emph{arXiv preprint arXiv:2005.00928}.

\bibitem[{Aky{\"u}rek et~al.(2023)Aky{\"u}rek, Schuurmans, Andreas, Ma, and Zhou}]{akyurek2023what}
Ekin Aky{\"u}rek, Dale Schuurmans, Jacob Andreas, Tengyu Ma, and Denny Zhou. 2023.
\newblock \href {https://openreview.net/forum?id=0g0X4H8yN4I} {What learning algorithm is in-context learning? investigations with linear models}.
\newblock In \emph{The Eleventh International Conference on Learning Representations}.

\bibitem[{Bengio et~al.(2009)Bengio, Louradour, Collobert, and Weston}]{bengio.etal_2009}
Yoshua Bengio, Jérôme Louradour, Ronan Collobert, and Jason Weston. 2009.
\newblock \href {https://doi.org/10.1145/1553374.1553380} {Curriculum learning}.
\newblock In \emph{Proceedings of the 26th {Annual} {International} {Conference} on {Machine} {Learning}}, {ICML} '09, pages 41--48, New York, NY, USA. Association for Computing Machinery.

\bibitem[{Besta et~al.(2023)Besta, Blach, Kubicek, Gerstenberger, Gianinazzi, Gajda, Lehmann, Podstawski, Niewiadomski, Nyczyk et~al.}]{besta2023graph}
Maciej Besta, Nils Blach, Ales Kubicek, Robert Gerstenberger, Lukas Gianinazzi, Joanna Gajda, Tomasz Lehmann, Michal Podstawski, Hubert Niewiadomski, Piotr Nyczyk, et~al. 2023.
\newblock Graph of thoughts: Solving elaborate problems with large language models.
\newblock \emph{arXiv preprint arXiv:2308.09687}.

\bibitem[{Brown et~al.(2020)Brown, Mann, Ryder, Subbiah, Kaplan, Dhariwal, Neelakantan, Shyam, Sastry, Askell, Agarwal, Herbert-Voss, Krueger, Henighan, Child, Ramesh, Ziegler, Wu, Winter, Hesse, Chen, Sigler, Litwin, Gray, Chess, Clark, Berner, McCandlish, Radford, Sutskever, and Amodei}]{brown2020language}
Tom~B. Brown, Benjamin Mann, Nick Ryder, Melanie Subbiah, Jared Kaplan, Prafulla Dhariwal, Arvind Neelakantan, Pranav Shyam, Girish Sastry, Amanda Askell, Sandhini Agarwal, Ariel Herbert-Voss, Gretchen Krueger, Tom Henighan, Rewon Child, Aditya Ramesh, Daniel~M. Ziegler, Jeffrey Wu, Clemens Winter, Christopher Hesse, Mark Chen, Eric Sigler, Mateusz Litwin, Scott Gray, Benjamin Chess, Jack Clark, Christopher Berner, Sam McCandlish, Alec Radford, Ilya Sutskever, and Dario Amodei. 2020.
\newblock \href {http://arxiv.org/abs/2005.14165} {Language models are few-shot learners}.

\bibitem[{{Center for High Throughput Computing}(2006)}]{chtc}
{Center for High Throughput Computing}. 2006.
\newblock \href {https://doi.org/10.21231/GNT1-HW21} {Center for high throughput computing}.

\bibitem[{Clark et~al.(2019)Clark, Khandelwal, Levy, and Manning}]{clark-etal-2019-bert}
Kevin Clark, Urvashi Khandelwal, Omer Levy, and Christopher~D. Manning. 2019.
\newblock \href {https://doi.org/10.18653/v1/W19-4828} {What does {BERT} look at? an analysis of {BERT}{'}s attention}.
\newblock In \emph{Proceedings of the 2019 ACL Workshop BlackboxNLP: Analyzing and Interpreting Neural Networks for NLP}, pages 276--286, Florence, Italy. Association for Computational Linguistics.

\bibitem[{Crawshaw(2020)}]{crawshaw2020multitask}
Michael Crawshaw. 2020.
\newblock \href {http://arxiv.org/abs/2009.09796} {Multi-task learning with deep neural networks: A survey}.

\bibitem[{Dong et~al.(2023)Dong, Li, Dai, Zheng, Wu, Chang, Sun, Xu, Li, and Sui}]{dong2023survey}
Qingxiu Dong, Lei Li, Damai Dai, Ce~Zheng, Zhiyong Wu, Baobao Chang, Xu~Sun, Jingjing Xu, Lei Li, and Zhifang Sui. 2023.
\newblock \href {http://arxiv.org/abs/2301.00234} {A survey on in-context learning}.

\bibitem[{Elhage et~al.(2021)Elhage, Nanda, Olsson, Henighan, Joseph, Mann, Askell, Bai, Chen, Conerly, DasSarma, Drain, Ganguli, Hatfield-Dodds, Hernandez, Jones, Kernion, Lovitt, Ndousse, Amodei, Brown, Clark, Kaplan, McCandlish, and Olah}]{elhage2021mathematical}
Nelson Elhage, Neel Nanda, Catherine Olsson, Tom Henighan, Nicholas Joseph, Ben Mann, Amanda Askell, Yuntao Bai, Anna Chen, Tom Conerly, Nova DasSarma, Dawn Drain, Deep Ganguli, Zac Hatfield-Dodds, Danny Hernandez, Andy Jones, Jackson Kernion, Liane Lovitt, Kamal Ndousse, Dario Amodei, Tom Brown, Jack Clark, Jared Kaplan, Sam McCandlish, and Chris Olah. 2021.
\newblock A mathematical framework for transformer circuits.
\newblock \emph{Transformer Circuits Thread}.
\newblock Https://transformer-circuits.pub/2021/framework/index.html.

\bibitem[{Garg et~al.(2022)Garg, Tsipras, Liang, and Valiant}]{garg2022icl}
Shivam Garg, Dimitris Tsipras, Percy~S Liang, and Gregory Valiant. 2022.
\newblock \href {https://proceedings.neurips.cc/paper_files/paper/2022/file/c529dba08a146ea8d6cf715ae8930cbe-Paper-Conference.pdf} {What can transformers learn in-context? a case study of simple function classes}.
\newblock In \emph{Advances in Neural Information Processing Systems}, volume~35, pages 30583--30598. Curran Associates, Inc.

\bibitem[{Graves et~al.(2017)Graves, Bellemare, Menick, Munos, and Kavukcuoglu}]{graves.etal_2017}
Alex Graves, Marc~G. Bellemare, Jacob Menick, Rémi Munos, and Koray Kavukcuoglu. 2017.
\newblock \href {https://proceedings.mlr.press/v70/graves17a.html} {Automated {Curriculum} {Learning} for {Neural} {Networks}}.
\newblock In \emph{Proceedings of the 34th {International} {Conference} on {Machine} {Learning}}, pages 1311--1320. PMLR.
\newblock ISSN: 2640-3498.

\bibitem[{Li et~al.(2023{\natexlab{a}})Li, Ildiz, Papailiopoulos, and Oymak}]{pmlr-v202-li23l}
Yingcong Li, Muhammed~Emrullah Ildiz, Dimitris Papailiopoulos, and Samet Oymak. 2023{\natexlab{a}}.
\newblock \href {https://proceedings.mlr.press/v202/li23l.html} {Transformers as algorithms: Generalization and stability in in-context learning}.
\newblock In \emph{Proceedings of the 40th International Conference on Machine Learning}, volume 202 of \emph{Proceedings of Machine Learning Research}, pages 19565--19594. PMLR.

\bibitem[{Li et~al.(2023{\natexlab{b}})Li, Ildiz, Papailiopoulos, and Oymak}]{li2023transformers}
Yingcong Li, Muhammed~Emrullah Ildiz, Dimitris Papailiopoulos, and Samet Oymak. 2023{\natexlab{b}}.
\newblock Transformers as algorithms: Generalization and stability in in-context learning.
\newblock In \emph{International Conference on Machine Learning}, pages 19565--19594. PMLR.

\bibitem[{Liu et~al.(2023)Liu, Yuan, Fu, Jiang, Hayashi, and Neubig}]{liu2023prompt}
Pengfei Liu, Weizhe Yuan, Jinlan Fu, Zhengbao Jiang, Hiroaki Hayashi, and Graham Neubig. 2023.
\newblock Pre-train, {Prompt}, and {Predict}: {A} {Systematic} {Survey} of {Prompting} {Methods} in {Natural} {Language} {Processing}.
\newblock \emph{ACM Computing Surveys}, 55.

\bibitem[{Lu et~al.(2023)Lu, Bigoulaeva, Sachdeva, Madabushi, and Gurevych}]{lu2023emergent}
Sheng Lu, Irina Bigoulaeva, Rachneet Sachdeva, Harish~Tayyar Madabushi, and Iryna Gurevych. 2023.
\newblock \href {http://arxiv.org/abs/2309.01809} {Are emergent abilities in large language models just in-context learning?}

\bibitem[{Min et~al.(2022{\natexlab{a}})Min, Lewis, Zettlemoyer, and Hajishirzi}]{min-etal-2022-metaicl}
Sewon Min, Mike Lewis, Luke Zettlemoyer, and Hannaneh Hajishirzi. 2022{\natexlab{a}}.
\newblock \href {https://doi.org/10.18653/v1/2022.naacl-main.201} {{M}eta{ICL}: Learning to learn in context}.
\newblock In \emph{Proceedings of the 2022 Conference of the North American Chapter of the Association for Computational Linguistics: Human Language Technologies}, pages 2791--2809, Seattle, United States. Association for Computational Linguistics.

\bibitem[{Min et~al.(2022{\natexlab{b}})Min, Lyu, Holtzman, Artetxe, Lewis, Hajishirzi, and Zettlemoyer}]{min-etal-2022-rethinking}
Sewon Min, Xinxi Lyu, Ari Holtzman, Mikel Artetxe, Mike Lewis, Hannaneh Hajishirzi, and Luke Zettlemoyer. 2022{\natexlab{b}}.
\newblock \href {https://doi.org/10.18653/v1/2022.emnlp-main.759} {Rethinking the role of demonstrations: What makes in-context learning work?}
\newblock In \emph{Proceedings of the 2022 Conference on Empirical Methods in Natural Language Processing}, pages 11048--11064, Abu Dhabi, United Arab Emirates. Association for Computational Linguistics.

\bibitem[{Olsson et~al.(2022)Olsson, Elhage, Nanda, Joseph, DasSarma, Henighan, Mann, Askell, Bai, Chen, Conerly, Drain, Ganguli, Hatfield-Dodds, Hernandez, Johnston, Jones, Kernion, Lovitt, Ndousse, Amodei, Brown, Clark, Kaplan, McCandlish, and Olah}]{olsson2022context}
Catherine Olsson, Nelson Elhage, Neel Nanda, Nicholas Joseph, Nova DasSarma, Tom Henighan, Ben Mann, Amanda Askell, Yuntao Bai, Anna Chen, Tom Conerly, Dawn Drain, Deep Ganguli, Zac Hatfield-Dodds, Danny Hernandez, Scott Johnston, Andy Jones, Jackson Kernion, Liane Lovitt, Kamal Ndousse, Dario Amodei, Tom Brown, Jack Clark, Jared Kaplan, Sam McCandlish, and Chris Olah. 2022.
\newblock In-context learning and induction heads.
\newblock \emph{Transformer Circuits Thread}.
\newblock Https://transformer-circuits.pub/2022/in-context-learning-and-induction-heads/index.html.

\bibitem[{Radford et~al.(2019)Radford, Wu, Child, Luan, Amodei, Sutskever et~al.}]{radford2019language}
Alec Radford, Jeffrey Wu, Rewon Child, David Luan, Dario Amodei, Ilya Sutskever, et~al. 2019.
\newblock Language models are unsupervised multitask learners.
\newblock \emph{OpenAI blog}, 1(8):9.

\bibitem[{Ruder(2017)}]{ruder2017overview}
Sebastian Ruder. 2017.
\newblock \href {http://arxiv.org/abs/1706.05098} {An overview of multi-task learning in deep neural networks}.

\bibitem[{Soviany et~al.(2022)Soviany, Ionescu, Rota, and Sebe}]{soviany.etal_2022}
Petru Soviany, Radu~Tudor Ionescu, Paolo Rota, and Nicu Sebe. 2022.
\newblock \href {https://doi.org/10.1007/s11263-022-01611-x} {Curriculum {Learning}: {A} {Survey}}.
\newblock \emph{International Journal of Computer Vision}, 130(6):1526--1565.

\bibitem[{Varshney et~al.(2022)Varshney, Mishra, and Baral}]{varshney.etal_2022}
Neeraj Varshney, Swaroop Mishra, and Chitta Baral. 2022.
\newblock \href {https://doi.org/10.18653/v1/2022.deeplo-1.13} {Let the {Model} {Decide} its {Curriculum} for {Multitask} {Learning}}.
\newblock In \emph{Proceedings of the {Third} {Workshop} on {Deep} {Learning} for {Low}-{Resource} {Natural} {Language} {Processing}}, pages 117--125, Hybrid. Association for Computational Linguistics.

\bibitem[{Vaswani et~al.(2017)Vaswani, Shazeer, Parmar, Uszkoreit, Jones, Gomez, Kaiser, and Polosukhin}]{vaswani2017attention}
Ashish Vaswani, Noam Shazeer, Niki Parmar, Jakob Uszkoreit, Llion Jones, Aidan~N Gomez, {\L}ukasz Kaiser, and Illia Polosukhin. 2017.
\newblock Attention is all you need.
\newblock \emph{Advances in neural information processing systems}, 30.

\bibitem[{Vig and Belinkov(2019)}]{vig-belinkov-2019-analyzing}
Jesse Vig and Yonatan Belinkov. 2019.
\newblock \href {https://doi.org/10.18653/v1/W19-4808} {Analyzing the structure of attention in a transformer language model}.
\newblock In \emph{Proceedings of the 2019 ACL Workshop BlackboxNLP: Analyzing and Interpreting Neural Networks for NLP}, pages 63--76, Florence, Italy. Association for Computational Linguistics.

\bibitem[{Von~Oswald et~al.(2023)Von~Oswald, Niklasson, Randazzo, Sacramento, Mordvintsev, Zhmoginov, and Vladymyrov}]{pmlr-v202-von-oswald23a}
Johannes Von~Oswald, Eyvind Niklasson, Ettore Randazzo, Joao Sacramento, Alexander Mordvintsev, Andrey Zhmoginov, and Max Vladymyrov. 2023.
\newblock Transformers learn in-context by gradient descent.
\newblock In \emph{Proceedings of the 40th International Conference on Machine Learning}, volume 202 of \emph{Proceedings of Machine Learning Research}, pages 35151--35174. PMLR.

\bibitem[{Wang et~al.(2021)Wang, Chen, and Zhu}]{wang.etal_2021}
Xin Wang, Yudong Chen, and Wenwu Zhu. 2021.
\newblock \href {https://doi.org/10.1109/TPAMI.2021.3069908} {A {Survey} on {Curriculum} {Learning}}.
\newblock \emph{IEEE Transactions on Pattern Analysis and Machine Intelligence}, pages 1--1.

\bibitem[{Wei et~al.(2022{\natexlab{a}})Wei, Tay, Bommasani, Raffel, Zoph, Borgeaud, Yogatama, Bosma, Zhou, Metzler, Chi, Hashimoto, Vinyals, Liang, Dean, and Fedus}]{wei2022emergent}
Jason Wei, Yi~Tay, Rishi Bommasani, Colin Raffel, Barret Zoph, Sebastian Borgeaud, Dani Yogatama, Maarten Bosma, Denny Zhou, Donald Metzler, Ed~H. Chi, Tatsunori Hashimoto, Oriol Vinyals, Percy Liang, Jeff Dean, and William Fedus. 2022{\natexlab{a}}.
\newblock \href {http://arxiv.org/abs/2206.07682} {Emergent abilities of large language models}.

\bibitem[{Wei et~al.(2022{\natexlab{b}})Wei, Wang, Schuurmans, Bosma, Xia, Chi, Le, Zhou et~al.}]{wei2022chain}
Jason Wei, Xuezhi Wang, Dale Schuurmans, Maarten Bosma, Fei Xia, Ed~Chi, Quoc~V Le, Denny Zhou, et~al. 2022{\natexlab{b}}.
\newblock Chain-of-thought prompting elicits reasoning in large language models.
\newblock \emph{Advances in Neural Information Processing Systems}, 35:24824--24837.

\bibitem[{Wei et~al.(2023)Wei, Wei, Tay, Tran, Webson, Lu, Chen, Liu, Huang, Zhou, and Ma}]{wei2023larger}
Jerry Wei, Jason Wei, Yi~Tay, Dustin Tran, Albert Webson, Yifeng Lu, Xinyun Chen, Hanxiao Liu, Da~Huang, Denny Zhou, and Tengyu Ma. 2023.
\newblock \href {http://arxiv.org/abs/2303.03846} {Larger language models do in-context learning differently}.

\bibitem[{Weiss et~al.(2016)Weiss, Khoshgoftaar, and Wang}]{weiss.etal_2016}
Karl Weiss, Taghi~M. Khoshgoftaar, and DingDing Wang. 2016.
\newblock \href {https://doi.org/10.1186/s40537-016-0043-6} {A survey of transfer learning}.
\newblock \emph{Journal of Big Data}, 3(1):9.

\bibitem[{Wolf et~al.(2019)Wolf, Debut, Sanh, Chaumond, Delangue, Moi, Cistac, Rault, Louf, Funtowicz, and Brew}]{wolf2019huggingface}
Thomas Wolf, Lysandre Debut, Victor Sanh, Julien Chaumond, Clement Delangue, Anthony Moi, Pierric Cistac, Tim Rault, R{\'{e}}mi Louf, Morgan Funtowicz, and Jamie Brew. 2019.
\newblock \href {http://arxiv.org/abs/1910.03771} {Huggingface's transformers: State-of-the-art natural language processing}.
\newblock \emph{CoRR}, abs/1910.03771.

\bibitem[{Xie et~al.(2022)Xie, Raghunathan, Liang, and Ma}]{xie2022an}
Sang~Michael Xie, Aditi Raghunathan, Percy Liang, and Tengyu Ma. 2022.
\newblock \href {https://openreview.net/forum?id=RdJVFCHjUMI} {An explanation of in-context learning as implicit bayesian inference}.
\newblock In \emph{International Conference on Learning Representations}.

\bibitem[{Xu et~al.(2020)Xu, Zhang, Mao, Wang, Xie, and Zhang}]{xu.etal_2020}
Benfeng Xu, Licheng Zhang, Zhendong Mao, Quan Wang, Hongtao Xie, and Yongdong Zhang. 2020.
\newblock \href {https://doi.org/10.18653/v1/2020.acl-main.542} {Curriculum {Learning} for {Natural} {Language} {Understanding}}.
\newblock In \emph{Proceedings of the 58th {Annual} {Meeting} of the {Association} for {Computational} {Linguistics}}, pages 6095--6104, Online. Association for Computational Linguistics.

\bibitem[{Yadlowsky et~al.(2023)Yadlowsky, Doshi, and Tripuraneni}]{yadlowsky2023pretraining}
Steve Yadlowsky, Lyric Doshi, and Nilesh Tripuraneni. 2023.
\newblock \href {http://arxiv.org/abs/2311.00871} {Pretraining data mixtures enable narrow model selection capabilities in transformer models}.

\bibitem[{Yang et~al.(2023)Yang, Hui, Yang, Li, Huang, and Li}]{yang2023iterative}
Jiaxi Yang, Binyuan Hui, Min Yang, Binhua Li, Fei Huang, and Yongbin Li. 2023.
\newblock \href {http://arxiv.org/abs/2305.13016} {Iterative forward tuning boosts in-context learning in language models}.

\bibitem[{Yao et~al.(2023)Yao, Yu, Zhao, Shafran, Griffiths, Cao, and Narasimhan}]{yao2023tree}
Shunyu Yao, Dian Yu, Jeffrey Zhao, Izhak Shafran, Thomas~L Griffiths, Yuan Cao, and Karthik Narasimhan. 2023.
\newblock Tree of thoughts: Deliberate problem solving with large language models.
\newblock \emph{arXiv preprint arXiv:2305.10601}.

\bibitem[{Yin et~al.(2023)Yin, Vig, Laban, Joty, Xiong, and Wu}]{yin-etal-2023-read}
Fan Yin, Jesse Vig, Philippe Laban, Shafiq Joty, Caiming Xiong, and Chien-Sheng Wu. 2023.
\newblock \href {https://doi.org/10.18653/v1/2023.acl-long.172} {Did you read the instructions? rethinking the effectiveness of task definitions in instruction learning}.
\newblock In \emph{Proceedings of the 61st Annual Meeting of the Association for Computational Linguistics (Volume 1: Long Papers)}, pages 3063--3079, Toronto, Canada. Association for Computational Linguistics.

\bibitem[{Zhai et~al.(2023)Zhai, Tong, Li, Cai, Qu, Lee, and Ma}]{zhai2023investigating}
Yuexiang Zhai, Shengbang Tong, Xiao Li, Mu~Cai, Qing Qu, Yong~Jae Lee, and Yi~Ma. 2023.
\newblock \href {http://arxiv.org/abs/2309.10313} {Investigating the catastrophic forgetting in multimodal large language models}.

\bibitem[{Zhang et~al.(2023)Zhang, Yu, Yu, Guo, and Jiang}]{zhang2023survey}
Zhihan Zhang, Wenhao Yu, Mengxia Yu, Zhichun Guo, and Meng Jiang. 2023.
\newblock A {{Survey}} of {{Multi-task Learning}} in {{Natural Language Processing}}: {{Regarding Task Relatedness}} and {{Training Methods}}.
\newblock In \emph{Proceedings of the 17th {{Conference}} of the {{European Chapter}} of the {{Association}} for {{Computational Linguistics}}}.

\end{thebibliography}

\clearpage

\appendix
\onecolumn

\section*{Appendix}
\section{Experimental Settings} \label{experimental_settings}
We train on the GPT-2 \cite{radford2019language} model (22.4 million parameters) provided by HuggingFace \cite{wolf2019huggingface} \footnote{ \url{https://huggingface.co/docs/transformers/model_doc/gpt2}} with 12 heads, 8 layers, and an embedding size of 256 over 500,000 steps, where each batch size is 64. Each batch consists of 100 ($x_i$, $f(x_i)$) pairs (we expect higher order polynomials to require more in-context examples to converge). During training time each model is evaluated every 2,000 steps on a validation dataset of size 32,000. During test time each model is evaluated on 64 randomly selected examples. We train GPT-2 using a A100-SXM4-80GB provided by the \citet{chtc}.

\section{Distribution Learning} \label{distribution_learning}
In addition to different function classes, we explore training data generated from different distributions, given that recent literature has shown that these models do not perform well under distributional shifts \cite{garg2022icl,yadlowsky2023pretraining}. Particularly, we sample inputs $x_i$ from (i) Gaussian distributions, (ii) skewed Gaussian distributions (decaying eigenvalues), and (iii) student-t distributions (df = 4). Attention analysis (Supplementary Figure~\ref{suppfig:attention_analysis_dist_learning_baseline} and~\ref{suppfig:attention_analysis_dist_learning_curriculum}) and normalized MSE (Supplementary Figure~\ref{suppfig:model_performance_dist_learning_baseline} and~\ref{suppfig:model_performance_dist_learning_curriculum}) across tasks may be found for both single-task and curriculum models in the Supplementary Materials. 

\section{Instruction Prompting} \label{instuction_prompting}

We explore two sets of instruction prompting architectures: one-hot encoded vectors and preset instruction vectors. The goal of instruction prompting was to evaluate whether our objective could benefit from instruction prompting the way language translation or other NLP tasks do.

\subsection{One Hot Encoded Instruction Vector (OHEI)}

After generating our $(x_i,f(x_i))$ pairs, we append a single one hot encoded vector $p$ to the beginning of the sequence, with the one hot encoding corresponds to the ``task'':

\begin{align}\nonumber
p = \begin{cases}
    p_0 = 1 & \varphi = \varphi_1 \\
    p_1 = 1 & \varphi = \varphi_2 \\
    p_2 = 1 & \varphi =\varphi_3 \\
    \end{cases}
\end{align}

We then apply a linear transformation to transform the concatenation into the dimension, 256, of our Transformer.

\subsection{Preset Instruction Vector (PI)} 
After we use a linear transformation to transform our $(x_i,f(x_i))$ pairs to the input dimension, 256, of our Transformer we append a unique vector, $p \sim \mathcal{N}(0,I_d)$, that has been sampled from an isotropic Gaussian distribution. This  ``instruction vector'' remains constant throughout the training of all models, but remains different for each of the different tasks.

\subsection{Instruction Prompting Remains Unclear} \label{instruction_unclear}

Supplementary Figure~\ref{suppfig:instruction_comparison} shows the comparison of a mixed curriculum with no instruction prompting, to the two instruction prompting architectures listed above, evaluated over all function class tasks. Applications of the one hot encoded instruction (OHEI) vector to the mixed curriculum  causes minimal improvement, whereas application of the preset instruction (PI) vector to the mixed curriculum  worsens model performance in the quadratic and cubic function class evaluation during test time. We believe the former has minimal effect in performance as the one-hot encoded vectors may just be seen as noise, whereas the latter most likely worsens the ability of the model to learn the task as it may be seen as an extreme version of noise (it may disrupt the flow of $x_i,f(x_i)$, confusing the model). Overall, we believe that instruction tokens may not be tractable in this setting due to the difficulty of learning a 20-dimensional instruction.

\onecolumn

\clearpage
\setcounter{figure}{0}



\section*{Supplementary Materials}
\begin{figure*}[ht]
    \centering
    \includegraphics[width=\textwidth]{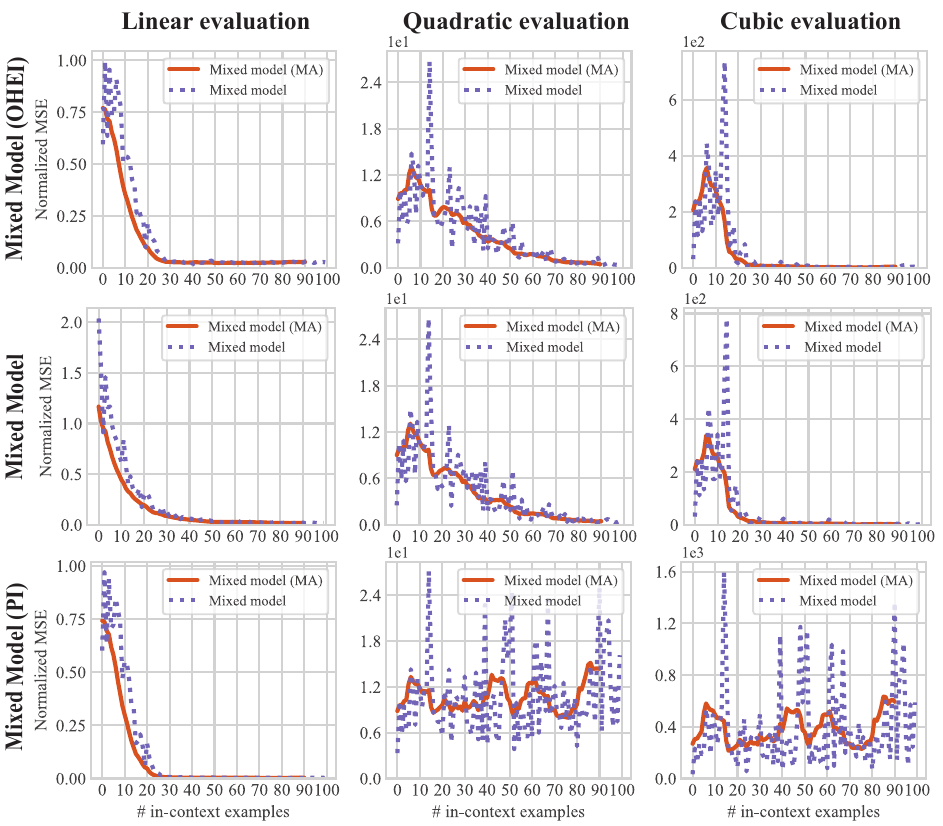}
    \vspace{0.05in}
    \caption[Model Performance - Comparison of different instruction prompting strategies in mixed curriculum model]{Normalized MSE over the number of in-context examples for the mixed curriculum model, mixed curriculum model with one hot encoded instruction (OHEI) vector and mixed curriculum model with preset instruction (PI) vector. Solid line represents the moving average (window = 10) whereas the dashed line is the true value. Scientific notation is used for the y-axis. Both of our attempts at instruction prompting are unsuccessful as normalized MSE remains the same or worsens across all tasks.}
     \label{suppfig:instruction_comparison}
\end{figure*}

\newpage

\begin{figure*}[ht]
    \centering
    \includegraphics[width=\textwidth]{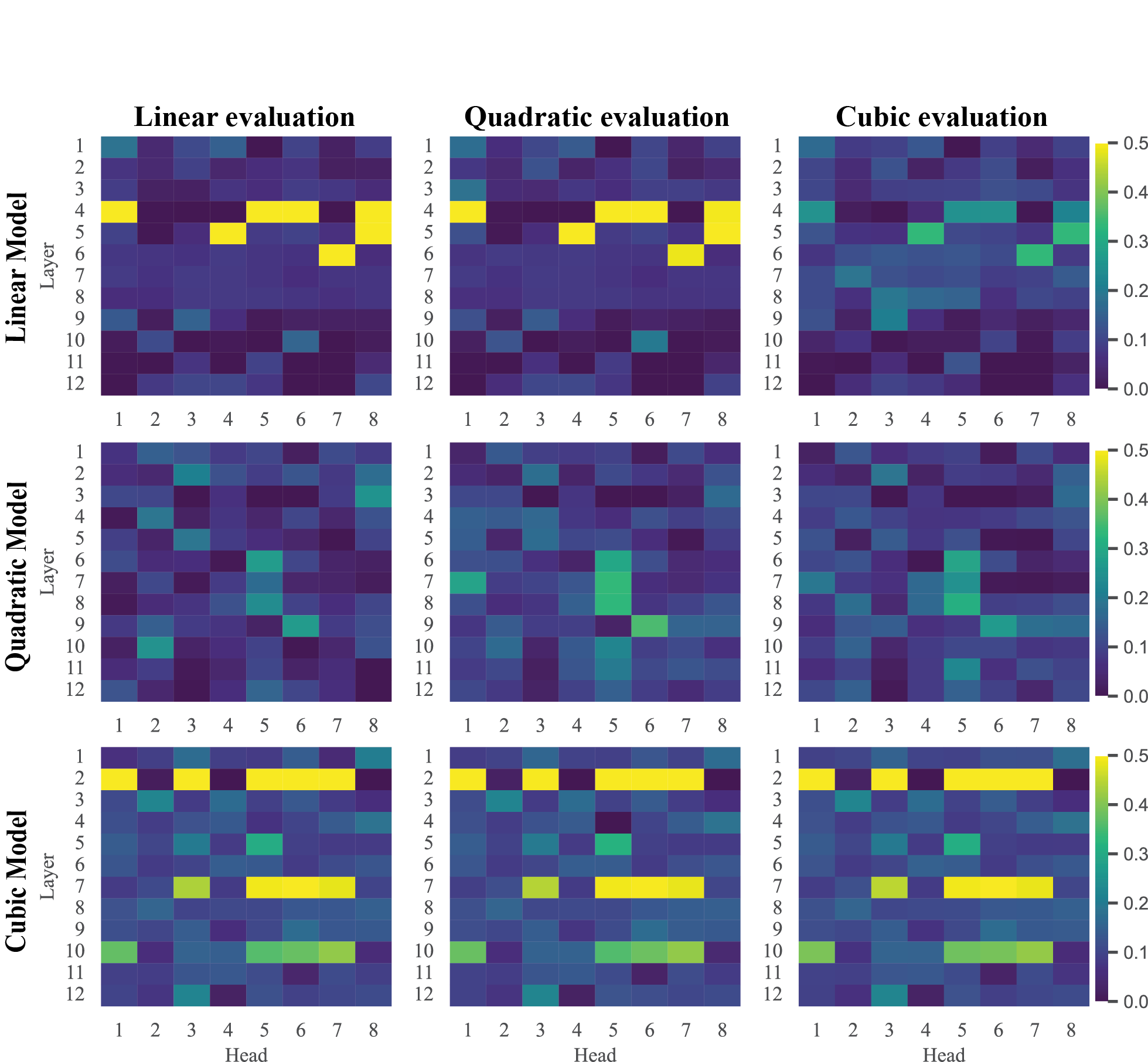}
    \vspace{0.05in}
    \caption[Attention Analysis - Single-Task Function Learning Models]{Attention analysis as described in Section \Scref{attention_analysis} for the single-task function class learning models. The linear model has different attention patterns when evaluated on the linear and cubic test time dataset as it has not seen cubic examples during training. The quadratic model has no retrospective heads as it does not converge, a fact that is made clear when analyzing normalized MSE in Supplementary Figure ~\ref{suppfig:model_performance_func_learning_baseline}. The cubic model seems to have learned the easier tasks (e.g. linear and quadratic) from learning the harder task (cubic).}
     \label{suppfig:attention_analysis_func_learning_baseline}
\end{figure*}

\newpage

\begin{figure*}[ht]
    \centering
    \includegraphics[width=\textwidth]{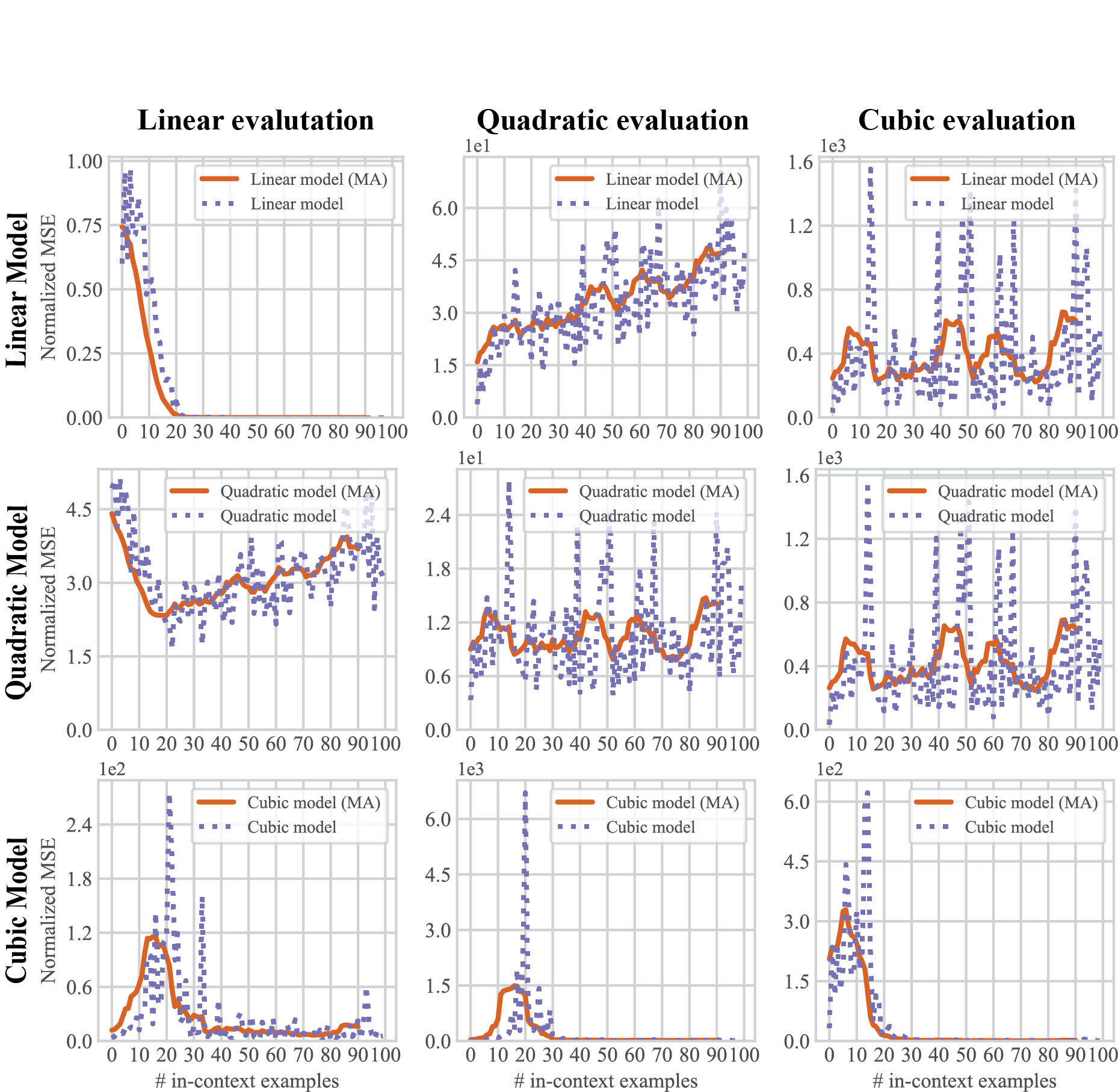}
    \vspace{0.05in}
    \caption[Model Performance - Single-Task Function Learning Models]{Normalized MSE over the number of in-context examples for the single-task function class learning models. Solid line represents the moving average (window = 10) whereas the dashed line is the true value. Scientific notation is used for the y-axis. The linear model is only able to achieve optimal MSE in the linear test time evaluation. The quadratic model never converges as it does not achieve optimal MSE in the quadratic test time evaluation, and as a result, does not perform well in the other task evaluation. The cubic model is able to achieve optimal MSE in the quadratic and cubic test time evaluation, however it struggles to perform well in the linear test time evaluation.}
     \label{suppfig:model_performance_func_learning_baseline}
\end{figure*}

\newpage

\begin{figure*}[ht]
    \centering
    \includegraphics[width=\textwidth]{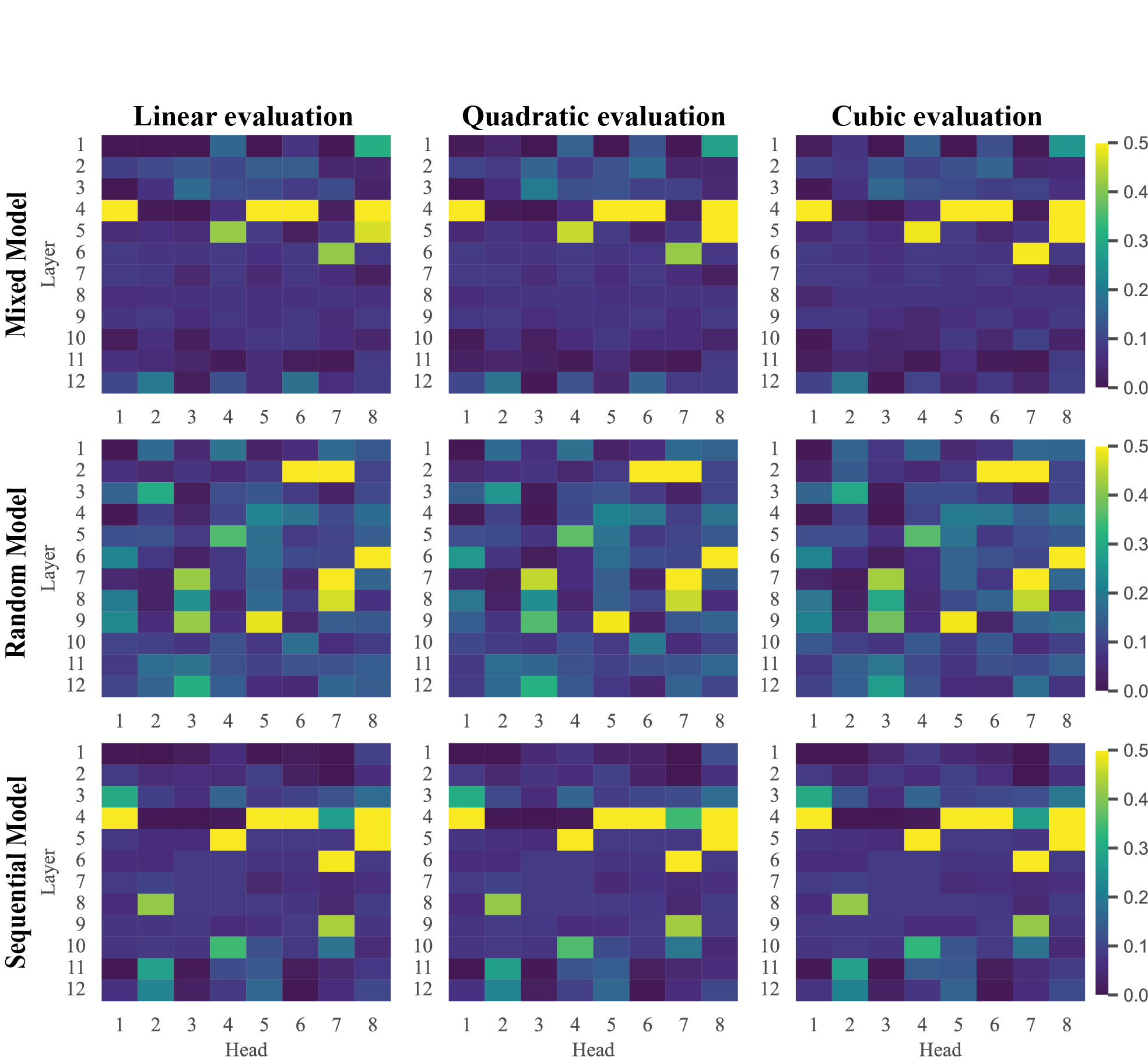}
    \vspace{0.05in}
    \caption[Attention Analysis - Curriculum Function Learning Models]{Attention analysis as described in Section \Scref{attention_analysis} for the curriculum function class learning models. All curriculum models maintain the same retrospective heads across tasks.}
     \label{suppfig:attention_analysis_func_learning_curriculum}
\end{figure*}

\newpage

\begin{figure*}[ht]
    \centering
    \includegraphics[width=\textwidth]{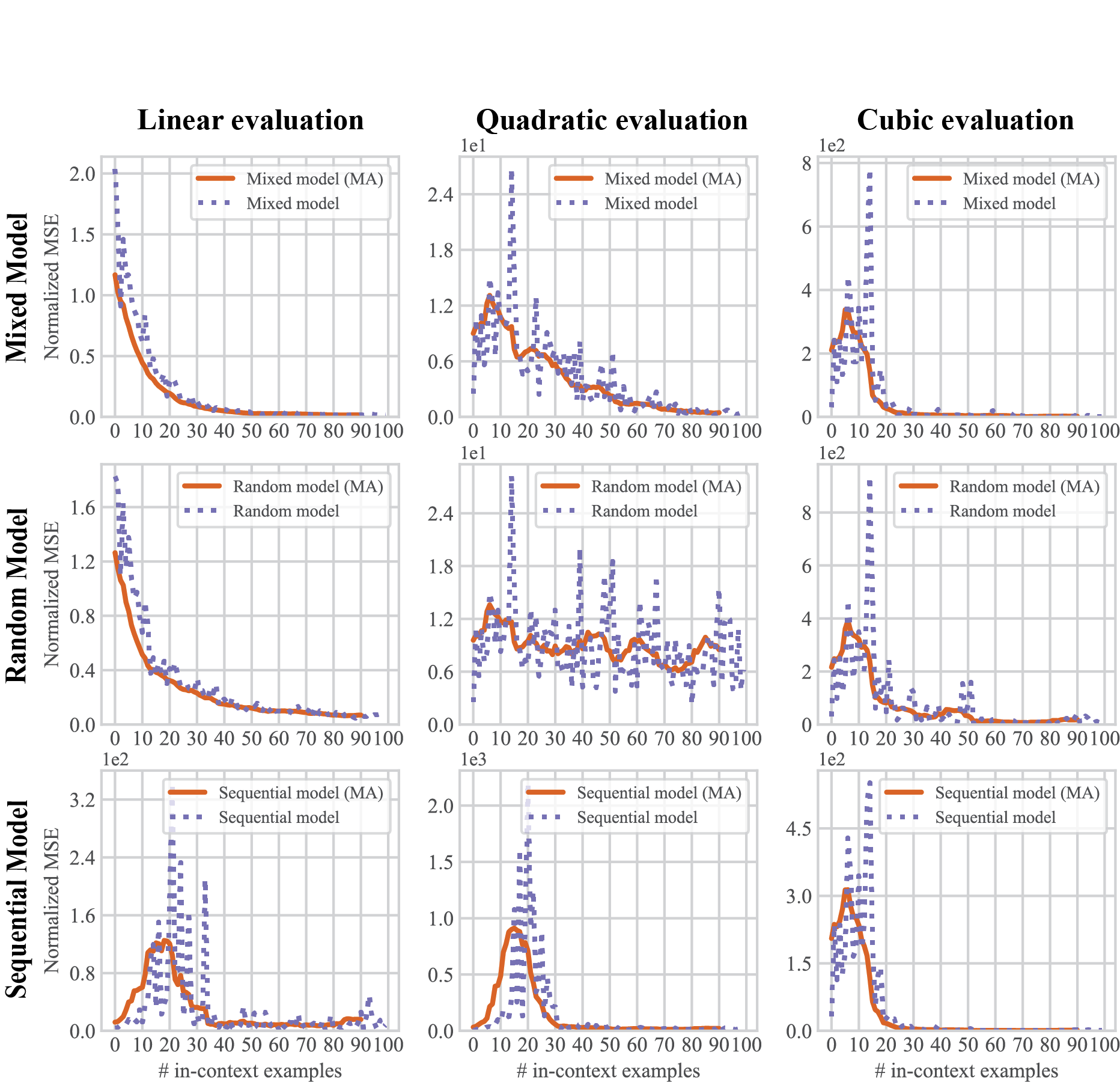}
    \vspace{0.05in}
    \caption[Model Performance - Curriculum Function Learning Models]{Normalized MSE over the number of in-context examples for the curriculum function class learning models. Solid line represents the moving average (window = 10) whereas the dashed line is the true value. Scientific notation is used for the y-axis. The mixed curriculum model outperforms the other curriculum models on all tasks. The random curriculum model performs well compared to the mixed curriculum model in linear and cubic evaluation, however it is unable to learn the quadratic function class. The sequential curriculum model is unable to learn any of the tasks.}
     \label{suppfig:model_performance_func_learning_curriculum}
\end{figure*}

\newpage


\begin{figure*}[ht]
    \centering
    \includegraphics[width=\textwidth]{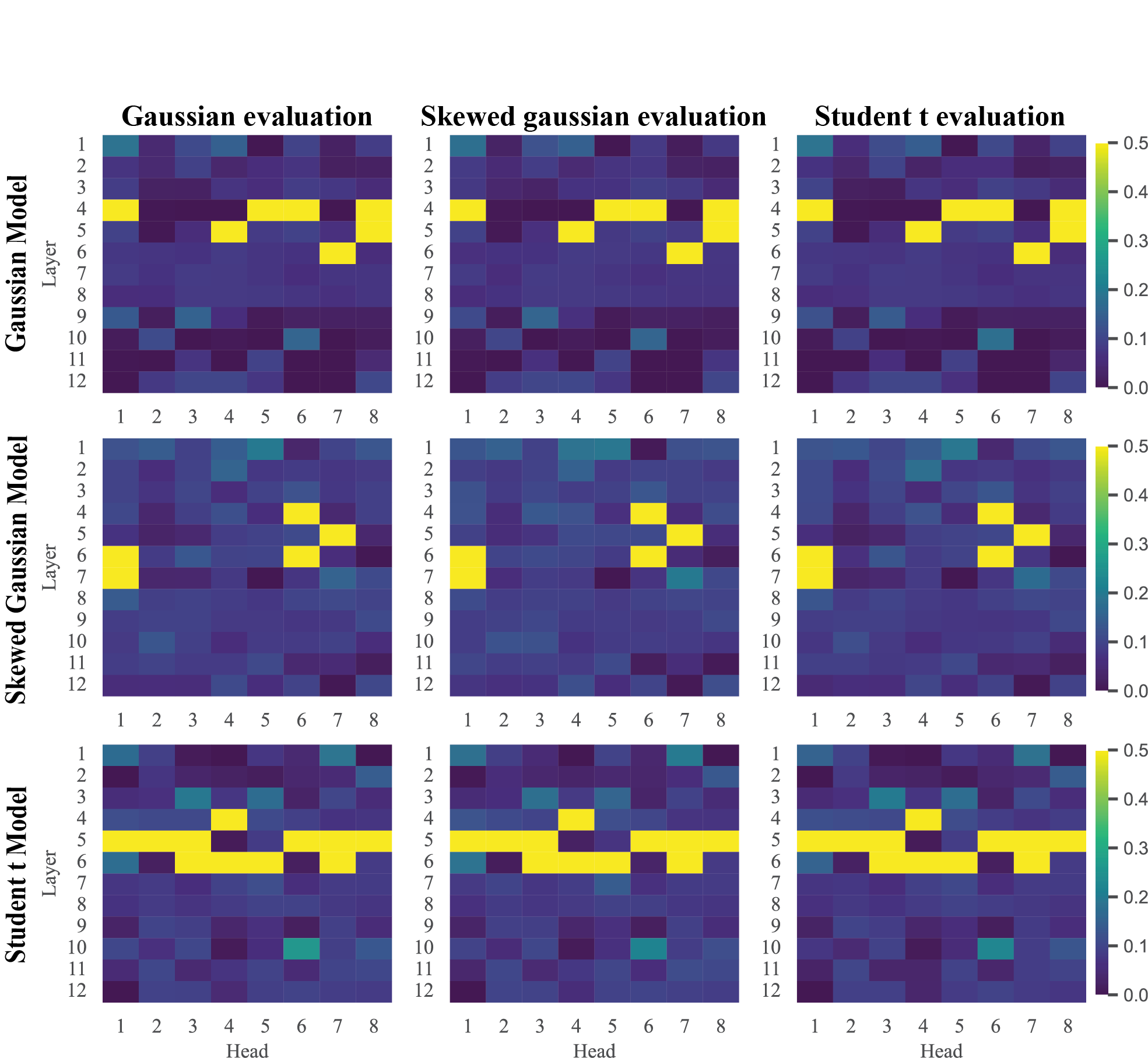}
    \vspace{0.05in}
    \caption[Attention Analysis - Single-Task Distribution Learning Models]{Attention analysis  as described in Section \Scref{attention_analysis} for the single-task distribution learning models. All single-task models keep the same retrospective heads across tasks. We hypothesize that this happens in this task and not function class learning as the $f(x_i)$ for the different distributions will be on a similar scale, which is not true in function class learning (e.g., linear will result in much smaller output than cubic).}
     \label{suppfig:attention_analysis_dist_learning_baseline}
\end{figure*}

\newpage

\begin{figure*}[ht]
    \centering
    \includegraphics[width=\textwidth]{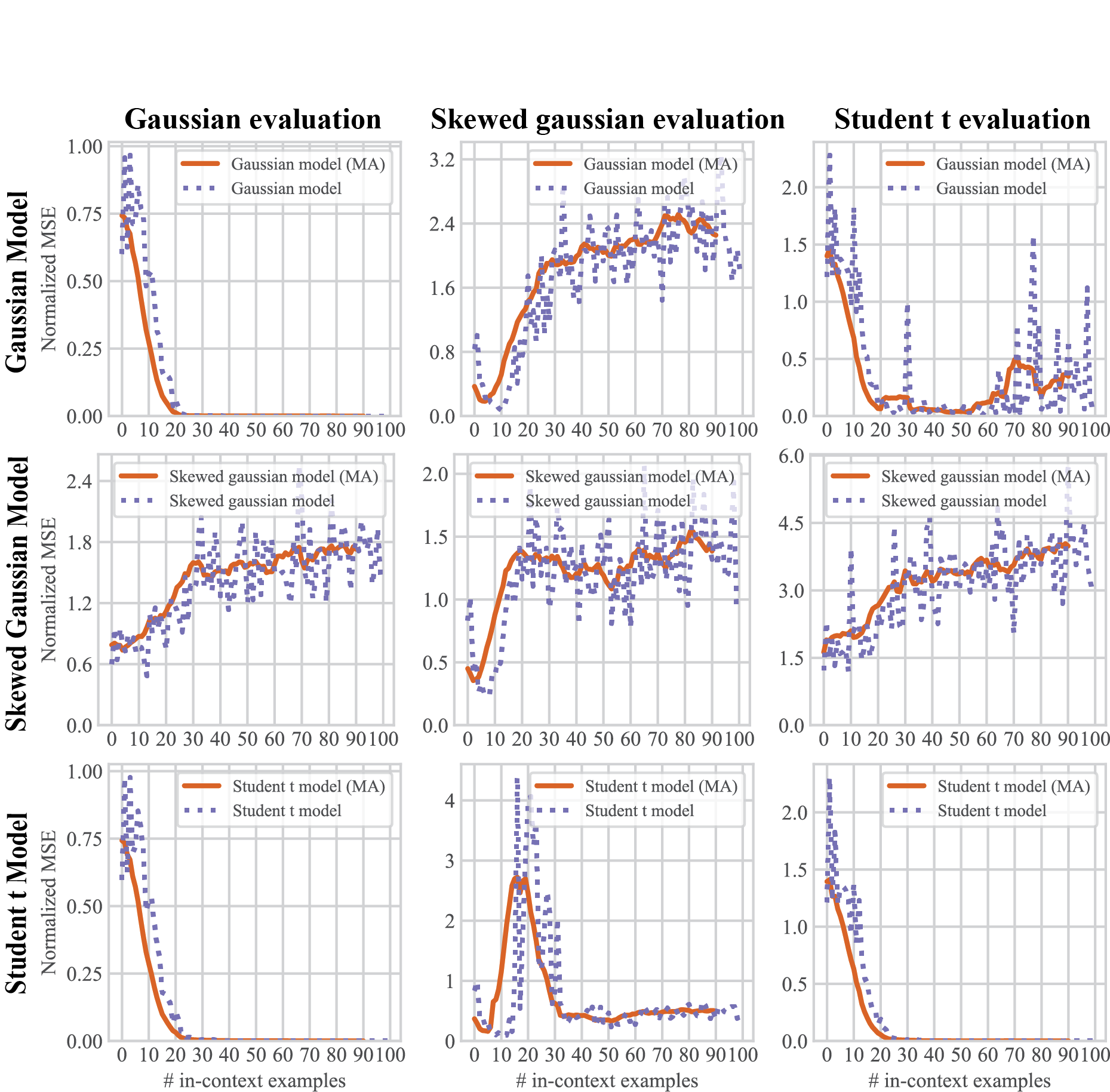}
    \vspace{0.05in}
    \caption[Model Performance - Single-Task Distribution Learning Models]{Normalized MSE over the number of in-context examples for the single-task distribution learning models. Solid line represents the moving average (window = 10) whereas the dashed line is the true value. Scientific notation is used for the y-axis. The Gaussian model may only be able to learn the Gaussian distribution as the skewed Gaussian and Student t distributions have tails that are too large. The skewed Gaussian model is unable to converge as indicated by the test time evaluation, resulting in poor performance in other tasks. The Student t model is able to learn all tasks relatively well as it's tailing is not as heavy as the skewed Gaussian distribution so it's able to learn the task.}
     \label{suppfig:model_performance_dist_learning_baseline}
\end{figure*}

\newpage

\begin{figure*}[ht]
    \centering
    \includegraphics[width=\textwidth]{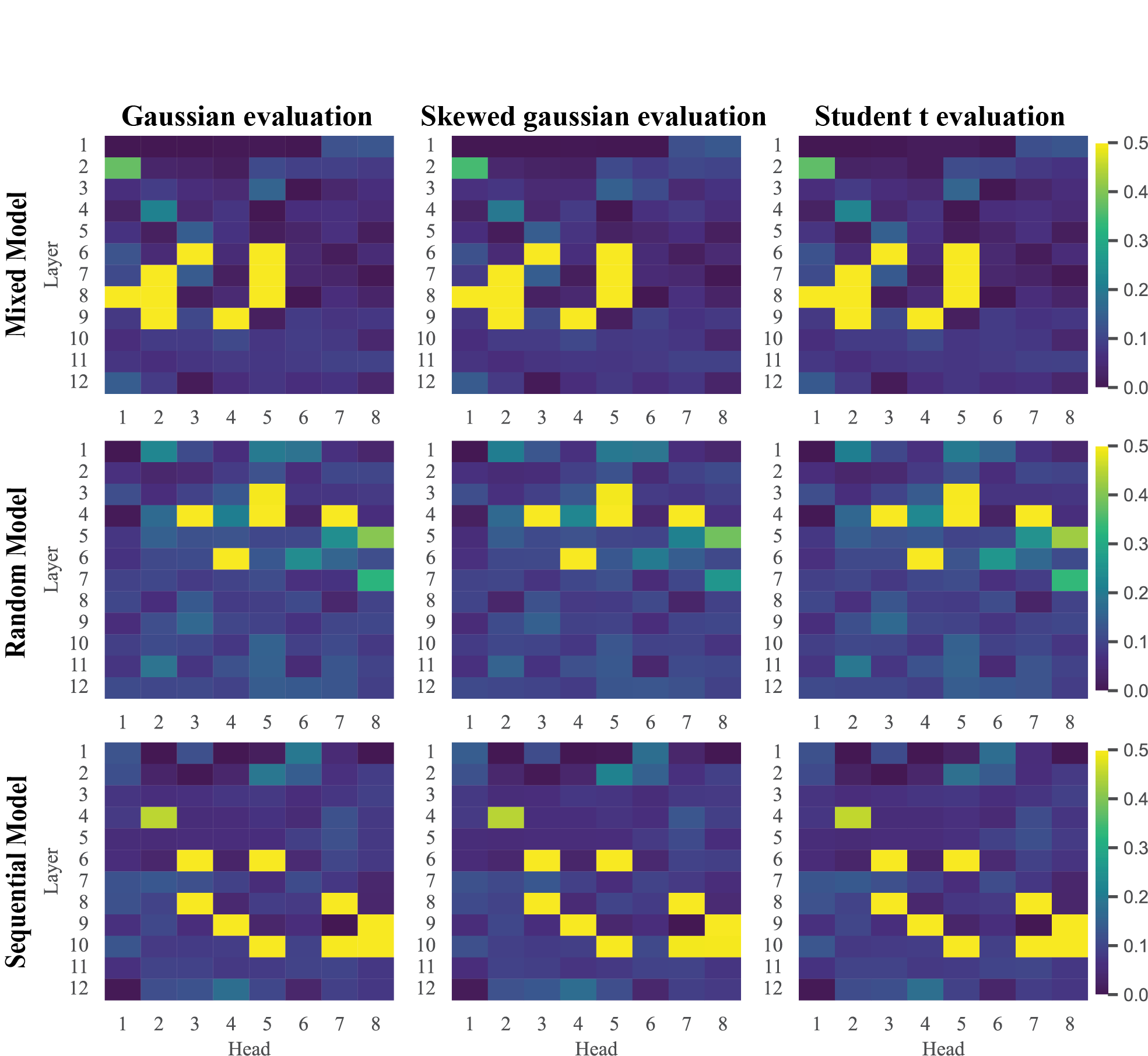}
    \vspace{0.05in}
    \caption[Attention Analysis - Curriculum Distribution Learning Models]{Attention analysis  as described in Section \Scref{attention_analysis} for the curriculum distribution learning models. All curriculum models maintain the same retrospective heads across tasks.}
     \label{suppfig:attention_analysis_dist_learning_curriculum}
\end{figure*}

\newpage

\begin{figure*}[ht]
    \centering
    \includegraphics[width=\textwidth]{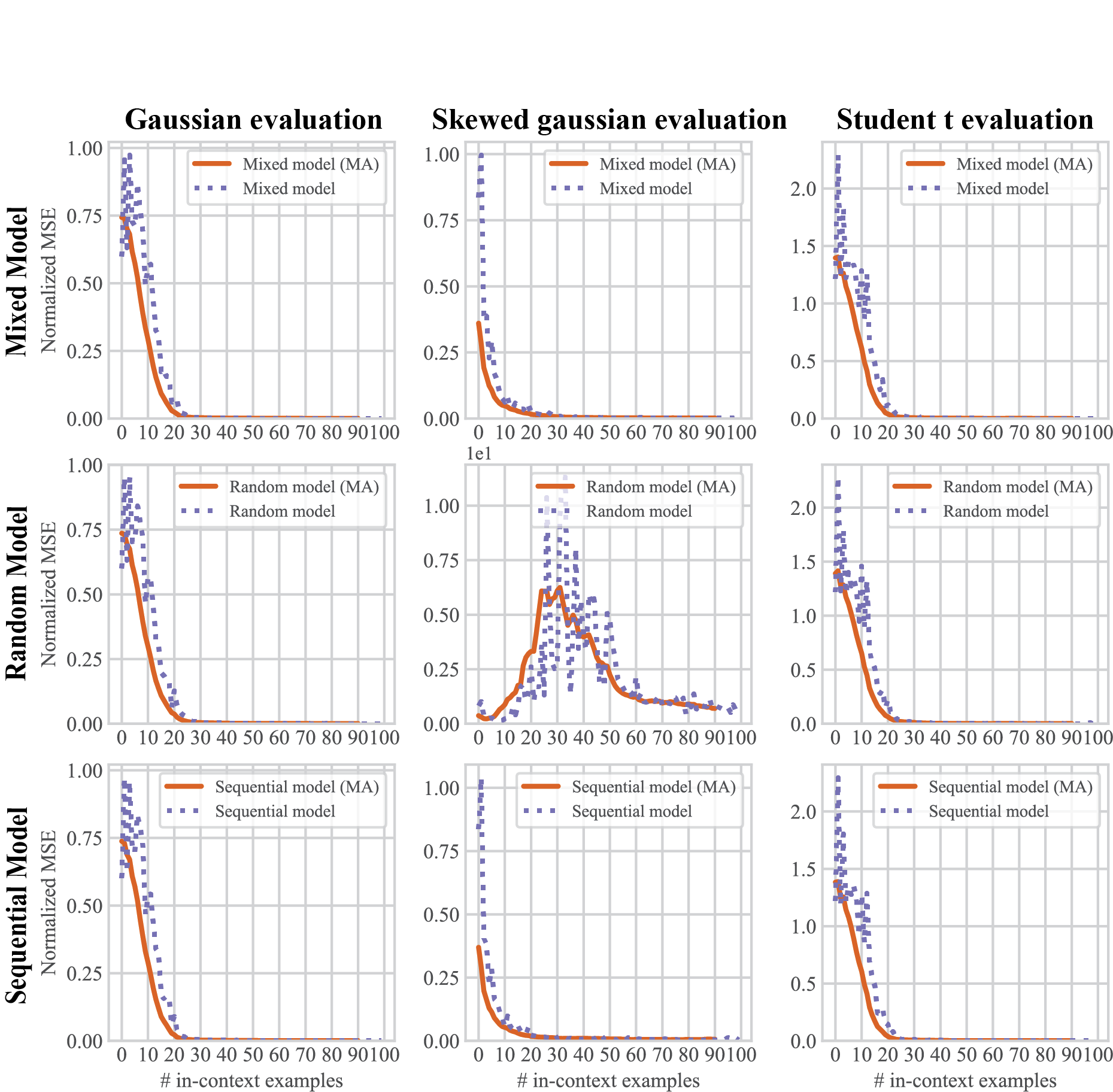}
    \vspace{0.05in}
    \caption[Model Performance - Curriculum Distribution Learning Models]{Normalized MSE over the number of in-context examples for the curriculum distribution learning models. Solid line represents the moving average (window = 10) whereas the dashed line is the true value. Scientific notation is used for the y-axis. All curriculum models seem to achieve optimal normalized MSE for all tasks, indicating that curriculum models are able to successfully learn several tasks.}
     \label{suppfig:model_performance_dist_learning_curriculum}
\end{figure*}

\end{document}